\documentclass[10pt,twocolumn,letterpaper]{article}
\usepackage{cvpr}              
\definecolor{cvprblue}{rgb}{0.21,0.49,0.74}
\usepackage[pagebackref,breaklinks,colorlinks,allcolors=cvprblue]
{hyperref}

\def\paperID{} 
\def\confName{CVPR}
\def\confYear{2026}

\title{
HTNav: A Hybrid Navigation Framework with Tiered Structure for Urban Aerial Vision-and-Language Navigation
}
\usepackage{multirow}
\usepackage{graphicx}
\usepackage{amsmath}
\usepackage{mathtools}
\usepackage{svg}

\renewcommand{\thefootnote}{\ifnum\value{footnote}=1 \textasteriskcentered\else\dag\fi}

\author{\textbf{Chengjie Fan} \thanks{These authors contributed equally.}  , \textbf{Cong Pan}  \footnotemark[\value{footnote}] , \textbf{Zijian Liu} ,  \textbf{Ningzhong Liu} ,  \textbf{Jie Qin} \thanks{Corresponding author.}
\\
Nanjing University of Aeronautics and Astronautics, China
\\
Key Laboratory of Brain-Machine Intelligence Technology, Ministry of Education, 
China
}

\begin{document}
\maketitle
\begin{abstract}
Inspired by the general Vision-and-Language Navigation (VLN) task, aerial VLN has attracted widespread attention, owing to its significant practical value in applications such as logistics delivery and urban inspection. However, existing methods face several challenges in complex urban environments, including insufficient generalization to unseen scenes, suboptimal performance in long-range path planning, and inadequate understanding of spatial continuity. To address these challenges, we propose HTNav, a new collaborative navigation framework that integrates Imitation Learning (IL) and Reinforcement Learning (RL) within a hybrid IL-RL framework. This framework adopts a staged training mechanism to ensure the stability of the basic navigation strategy while enhancing its environmental exploration capability. By integrating a tiered decision-making mechanism, it achieves collaborative interaction between macro-level path planning and fine-grained action control. Furthermore, a map representation learning module is introduced to deepen its understanding of spatial continuity in open domains. On the CityNav benchmark, our method achieves state-of-the-art performance across all scene levels and task difficulties. Experimental results demonstrate that this framework significantly improves navigation precision and robustness in complex urban environments.

\end{abstract}    
\section{Introduction}
In recent years, Vision-and-Language Navigation (VLN) has achieved remarkable progress. With continuous advancements in computer vision and natural language processing \cite{gaiaun, depth}, agents are now capable not only of understanding complex visual information, but also of navigating by following natural language instructions. Meanwhile, aerial VLN has attracted growing attention. Due to reduced constraints from ground terrain and flexible deployment, Unmanned Aerial Vehicles (UAVs) demonstrate significant potential for practical applications
in diverse urban environments, offering enhanced capabilities for efficient urban management, disaster monitoring, and related tasks.

\begin{figure}[!htb]
\centering
    \includegraphics[scale=0.40]{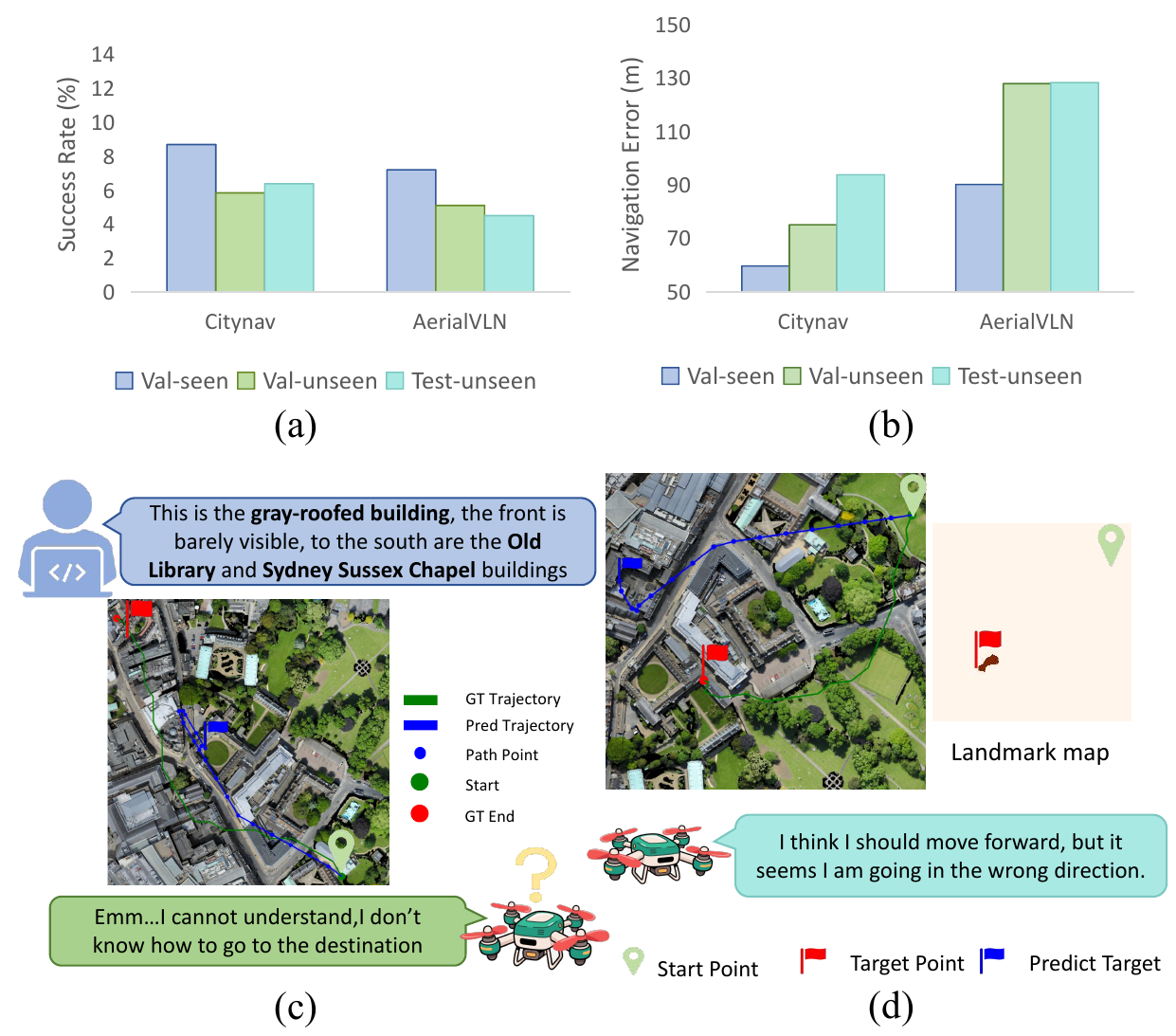}
\caption{
    Significant challenges include:
    (a–b) show decreased success rates and increased navigation errors when generalizing from seen to unseen scenes;
    (c) illustrates UAVs losing track of the goal in long-range navigation;
    (d) reveals directional errors caused by failing to interpret complex spatial information.
}
\label{figure1}
\end{figure}
With the emergence of the AVDN \cite{c:avdn} task, an increasing number of aerial VLN approaches have been proposed in recent years. For example, methods such as CityNav \cite{citynav}, FG-AVDN \cite{c:fgavdn}, and FlightGPT \cite{flightgpt} focus on realistic satellite remote sensing and urban scene imagery, using a top-down 2D perspective to generate UAV navigation paths based on given instructions. Meanwhile, methods like AerialVLN \cite{aerialvln}, OpenUAV \cite{openuav}, and OpenFLY \cite{openfly} are built upon simulation environments such as AirSim \cite{airsim}. They introduce diverse datasets for UAV navigation in 3D-modeled scenarios and offer a more immersive navigation experience. By integrating visual and linguistic information, these methods continuously advance the autonomous navigation capabilities of UAVs in complex environments.

Although the field of aerial VLN has made significant progress in recent years, UAV navigation in complex, real-world urban environments still faces substantial challenges. As demonstrated in Figure \ref{figure1}, these challenges manifest in several ways. First, existing methods \cite{aerialvln,stmr,citynav} exhibit notable limitations in generalization and environmental exploration, resulting in overall low success rates. This deficiency is especially pronounced in unseen scenarios, where the navigation failure rate increases significantly and navigation errors accumulate. Second, the performance of current methods on long-range navigation tasks is often unsatisfactory \cite{openfly,geonav}. Accumulated errors during the iterative decision-making process can easily cause UAVs to lose accurate localization, ultimately leading to navigation failure. Finally, a precise understanding of spatial information is central to UAV navigation,  requiring the system to accurately parse spatial relationships to plan a path from start to destination. However, current methods still have limitations in this regard \cite{citynav,stmr}.

To address the challenges of long-range UAV navigation in complex and unseen environments, we propose HTNav, a tiered cognitive navigation framework that integrates Imitation Learning (IL) and Reinforcement Learning (RL) into a hybrid IL-RL paradigm. Our primary innovation in the learning mechanism lies in a staged training strategy. Specifically, in the IL stage, the model learns a robust baseline policy from expert demonstrations. This policy is then further optimized in the RL stage through continuous interaction with the environment.

In terms of decision making, HTNav adopts a tiered cognitive architecture. At the macroscopic level, the system leverages landmarks, the UAV state, and semantic map features to reason and generate informative intermediate waypoints, thereby avoiding suboptimal local solutions that often arise in long-range navigation. At the microscopic level, the system makes fine-grained action decisions by integrating real-time observations with contextual information provided by the semantic map.

To efficiently and accurately represent spatial information, we design a map representation learning module based on a residual network \cite{resnet}. This module encodes the map using residual connections, effectively preserving fine-grained spatial details and local geometric continuity during feature extraction. To further enhance the discriminability of encoded features, we introduce an efficient SCConv module \cite{attention, scconv}, which adaptively identifies and suppresses redundant features in both the spatial and channel dimensions of feature maps. It improves the efficiency and selectivity of map feature utilization for downstream decision making.

Furthermore, we collaboratively conducted targeted refinements to the existing CityNav dataset, establishing a more robust benchmark for evaluation.

In summary, our contributions are as follows:
\begin{itemize}
\item We propose HTNav, a hybrid navigation architecture that employs a collaborative optimization mechanism integrating imitation learning and reinforcement learning, thereby enabling cognitively enhanced navigation with robust policies in complex urban environments.
\item A tiered cognitive decision-making mechanism facilitates multi-level control through task decoupling.
\item An innovative map representation learning module enables precise understanding of multi-source spatial semantics and geometric relationships in open domains.
\item Through targeted manual data refinement of the CityNav dataset, our approach demonstrates robust performance across both the original and revised versions of CityNav. Comprehensive experiments verify consistent superiority over all state-of-the-art baselines in key metrics.
\end{itemize}

\section{Related Work}
\subsection{Vision-and-Language Navigation}


\begin{figure*}[ht]

\begin{center}

\includegraphics[width=18cm,height=8.5cm]{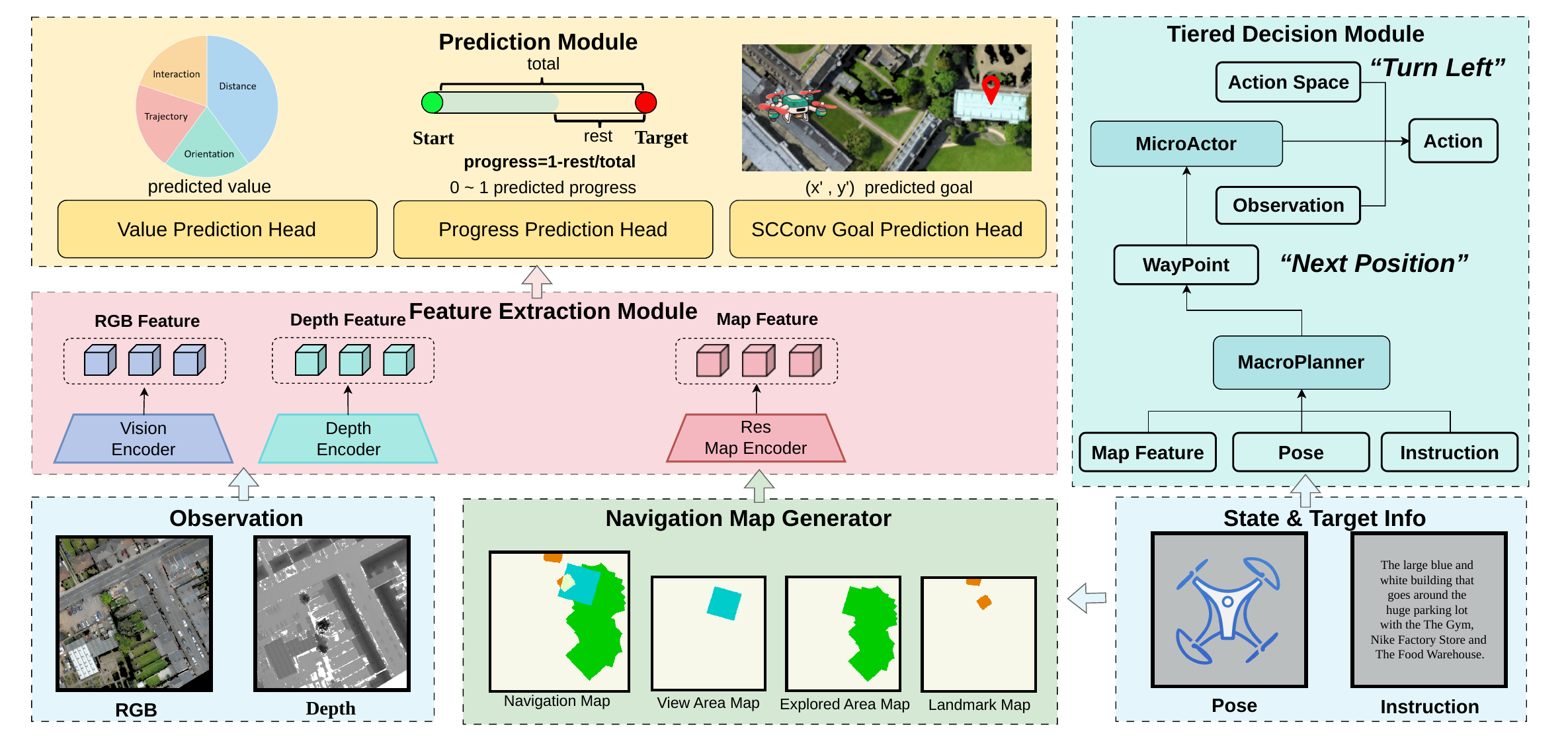}

\end{center}

\caption{
\textbf{Architecture of HTNav.} The model processes visual observations and generates a multi-layered navigation map from state and target information. Features extracted from RGB, depth, and map inputs by their respective encoders are fed into a three-head prediction module: a Value Prediction Head for expected cumulative reward, a Progress Prediction Head for navigation status, and a Goal Prediction Head for the final destination. In parallel, a decision module uses this information to generate waypoints and execute actions.
}

\label{figure2}

\end{figure*}

Ground-level VLN has progressed from early indoor instruction-following benchmarks toward longer horizons, multilingualism, and interactivity. R2R \cite{r2r} formulates navigation between panoramic viewpoints given natural language instructions. R4R \cite{r4r} extends trajectory and instruction length via path concatenation, while RxR \cite{rxr} adds multilingual coverage at scale and richer descriptions. \mbox{VLN-CE} \cite{vlnce} introduces continuous 3D control to reduce the simulation-to-real gap. REVERIE \cite{reverie} focuses on object grounding and localization. Embodied Question \mbox{Answering} (EQA) \cite{embodiedqa} enables dialog-based disambiguation, and ALFRED \cite{alfred} targets multi-step household manipulation. TouchDown \cite{touchdown} addresses outdoor urban challenges with street-level imagery and spatially grounded instructions. Landmark-RxR \cite{landmarkrxr} provides landmark-level annotations for fine-grained instruction–trajectory alignment, while FDA \cite{frequency} mixes high-frequency Fourier components to yield model-agnostic augmented views for R2R, RxR, and REVERIE. 
Collectively, these datasets and strategies constitute the ground-level VLN evaluation landscape.

\subsection{Aerial Vision-and-Language Navigation}
Aerial VLN has diversified from pioneering instruction-following datasets to benchmarks and frameworks emphasizing realistic flight dynamics, multi-view fusion, and urban-scale semantic grounding. AerialVLN \cite{aerialvln} introduces the first UAV-centric VLN benchmark for outdoor 3D instruction following. STMR \cite{stmr} improves aerial trajectory reliability via grid-based view selection and BEV map integration \cite{baeformer}. OpenUAV \cite{openuav} and OpenFLY \cite{openfly} offer greater motion freedom and automated toolchains, mitigating constraints imposed by discrete actions and simplified dynamics. NavAgent \cite{navagent} tackles small-landmark recognition with GLIP fine-tuning and dynamic graph encoding, and releases the Landmark2K dataset.
Beyond instruction following, AVDN \cite{c:avdn} establishes a vision-and-dialog navigation setting for UAVs; FG-AVDN \cite{c:fgavdn} scales annotation via a semi-automatic pipeline and introduces fine-grained entity–landmark alignment. CityNav \cite{citynav} defines city-scale aerial navigation; subsequent methods such as GeoNav \cite{geonav}, FlightGPT \cite{flightgpt}, and SA-GCS \cite{SA-GCS} decompose navigation into sequential subgoals using vision-language models. Recent studies show that RL fine-tuning improves generalization in aerial VLN. Our approach advances this paradigm with a novel, synergistic IL-RL framework. It pre-trains a state-value function during the imitation stage, using its weights to initialize the RL critic. This provides the agent with a high-quality value baseline, transforming its exploration from unguided to guided. Consequently, our method enhances generalization and significantly boosts exploration efficiency and stability.

\section{Method}
\subsection{Task Formulation}
The CityNav task requires a UAV to navigate to a specific target location described by a context-aware natural language instruction. At each timestep, the environment provides the UAV with a linguistic description of the goal along with the current pose of the UAV. During navigation, the UAV utilizes top-down RGB and depth images as real-time visual feedback, while simultaneously constructing a dynamic incremental navigation map based on its past trajectory to guide path planning.

The task environment leverages a pre-constructed landmark map containing geographic priors. When the language instruction refers to a specific landmark (e.g., "Nike Factory"), this map offers spatial references for localization. The UAV integrates these multimodal inputs to make sequential decisions and continues until it autonomously determines that it has reached the target area. A navigation episode is deemed successful if the distance from the final position to the ground-truth target is within 20 meters.
Building upon MGP \cite{2024citynav}, we propose the HTNav navigation framework, as illustrated in Figure~\ref{figure2}, to address the challenges of city navigation.

\subsection{Hybrid Navigation Architecture}

Unlike IL, which learns from pre-collected expert demonstrations, RL enables agents to explore optimal strategies through trial-and-error interactions with the environment in complex, observable settings \cite{large-scale, hybrid}. This mechanism allows RL to go beyond the limitations of these demonstrations by autonomously discovering novel solutions through exploration, thereby significantly improving the model’s novelty and applicability. Trained RL models also exhibit strong transferability to tasks with similar structures or objectives \cite{gaia}. However, RL suffers from substantial drawbacks, most notably its prolonged training process \cite{SA-GCS}.

To leverage their complementary strengths while mitigating their limitations, we propose a hybrid IL-RL strategy. IL first extracts features and generates approximate solutions from historical data, providing high-quality initialization and policy priors. Subsequently, RL is introduced to further optimize policies through continuous environmental exploration. This approach leverages the capacity of IL to rapidly acquire knowledge from static data, while simultaneously preserving the capability of RL to explore unknown policy spaces through environmental interaction, without being constrained to the demonstration data distribution.

The staged training framework operates as follows. In stage 1, we train a multi-task goal predictor using expert demonstration trajectories. This model incorporates an explicit value function head that simultaneously predicts: (a) targets including position and progress; and (b) the state-value function \(V(s_t)\). The state-value function \(V(s_t)\) estimates the expected discounted return from state \(s_t\):
\begin{equation}
V(s_t) = \mathbb{E} \left[ \sum_{k=0}^{\infty} \gamma^k r_{t+k} \mid s_t \right].
\end{equation}
Here, \(V(s_t)\) denotes the expected discounted return from state \(s_t\), where \(\gamma\) is the discount factor and \(r_{t+k}\) is the reward at time \(t+k\).
This encourages the model to capture the environment’s reward structure early in training.



In stage 2, we fine-tune the policy using the Proximal Policy Optimization (PPO) algorithm~\cite{ppo}. PPO is designed to ensure stable training by preventing excessively large policy updates. This is achieved through a clipped surrogate objective function:
\begin{equation}
L^{\text{CLIP}}(\theta) = \mathbb{E}_t \left[ \min \left( r_t(\theta) \hat{A}_t, \operatorname{clip}\left(r_t(\theta), 1-\epsilon, 1+\epsilon\right) \hat{A}_t \right) \right].
\end{equation}
Here, $r_t(\theta)$ is the probability ratio between the new policy and the old policy, $\hat{A}_t$ is the advantage estimate, and $\epsilon$ is the clipping hyperparameter. The objective function $L^{\text{CLIP}}(\theta)$ constrains the policy update by clipping the ratio $r_t(\theta)$, which enhances training stability. To further accelerate convergence, the PPO value network is initialized with the value function parameters learned in stage 1.



The total loss for our joint training framework is a weighted combination of imitation and reinforcement learning objectives. It is defined piecewise as:
\begin{equation}
L_{\text{total}} =
\begin{cases}
L_{\text{IL}} + L_{\text{V}} + \lambda_{\text{RL}} L_{\text{RL}}, & \text{if RL is enabled,} \\
L_{\text{IL}}, & \text{otherwise.}
\end{cases}
\end{equation}
Here, $\lambda_{\text{RL}} \in [0,1]$ is a coefficient balancing the contribution of the RL loss. The imitation loss $L_{\text{IL}}$ and the value function loss $L_{\text{V}}$ are both formulated using the MSE:
\begin{align}
L_{\text{IL}} &= \text{MSE}(\hat{g}, g) + \text{MSE}(\hat{p}, p). \\
L_{\text{V}} &= \text{MSE}(\hat{v}, v).
\end{align}
Here, $\hat{g}$ and $g$ denote the predicted and ground-truth goal positions. Similarly, $(\hat{p}, p)$ and $(\hat{v}, v)$ denote the predicted and ground-truth progress and state-value, respectively.

The reward function integrates four components to optimize navigation: 
a distance reward encouraging target approach, a direction reward promoting heading alignment, a goal reward triggered when the agent is within a threshold distance of the target, and a step penalty discouraging inefficient exploration. These components combine to form the raw reward at time $t$:
\begin{equation}
r_t^{\text{raw}} = \alpha(d_{t-1}-d_t) + \beta\left(1 - \frac{|\theta_t - \theta_t^*|}{\pi}\right) + \eta\,\mathbb{I}(d_t < d_{\text{goal}}) + \delta.
\end{equation}
Here, $\alpha$, $\beta$, and $\eta$ weight the first three terms; 
$\delta$ is the constant step penalty; 
$d_{t-1}$ and $d_t$ are the distances to the target at successive time steps; 
$\theta_t$ is the agent's current heading angle; 
$\theta_t^*$ is the desired heading toward the target. We compute $|\theta_t - \theta_t^*|$ as the wrapped angular difference in $[0,\pi]$.
$d_{\text{goal}}$ is the distance threshold that activates the goal reward via the indicator function $\mathbb{I}(\cdot)$. To ensure numerical stability during training, we clip the raw reward:
\begin{equation}
r_t = \operatorname{clip}\!\left(r_t^{\text{raw}},\, r_{\min},\, r_{\max}\right).
\end{equation}
where $r_{\min}$ and $r_{\max}$ specify the reward bounds.

\subsection{Tiered Decision Mechanism}
\begin{figure}[!h]

\begin{center}

\includegraphics[scale=0.60]{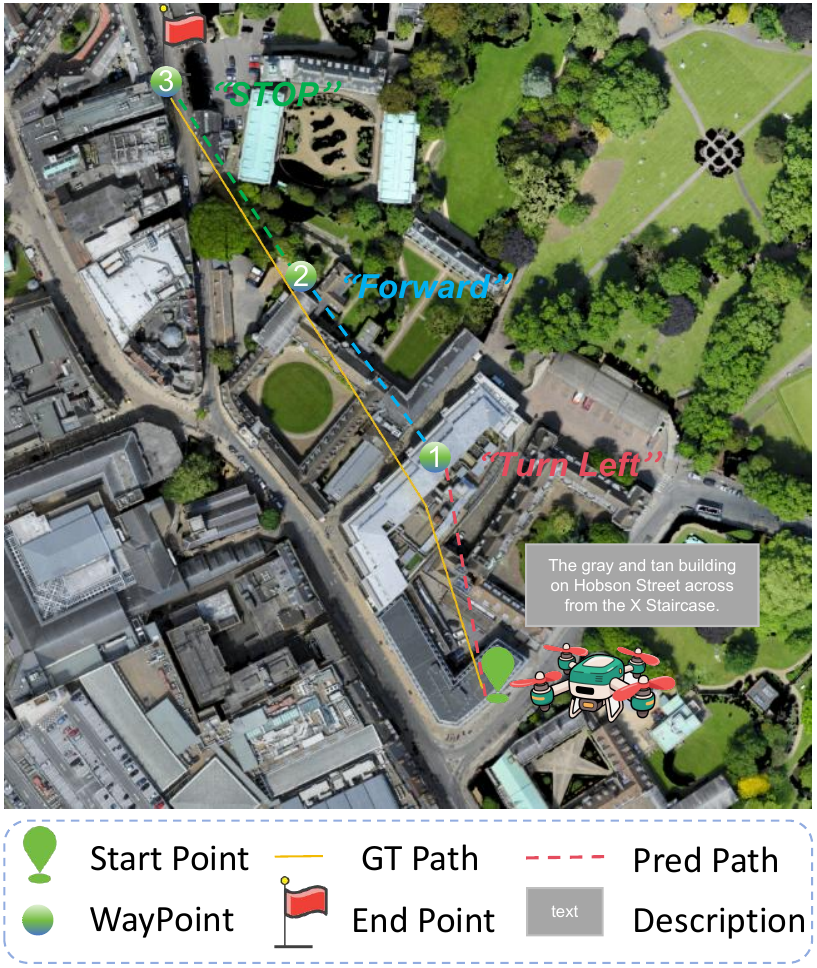}

\end{center}

\caption{A schematic of the Tiered Decision Mechanism. The plot shows the UAV's ground-truth and predicted paths, annotated with critical waypoints and action decisions (e.g. "Turn left").}

\label{figure5}

\end{figure}


To address the challenges of path planning and action selection in long-range UAV navigation, we propose a tiered decision-making mechanism. Unlike traditional methods such as the teacher algorithm in MGP, which are often myopic and brittle, these approaches rely on rigid precomputed paths and ignore real-time perception. In contrast, our architecture decomposes the task into a high-level MacroPlanner that ensures global path rationality and a low-level MicroActor that enables reactive local control. This strategic decomposition enhances overall navigational robustness and adaptability. Figure \ref{figure5} provides a visualization of a sample trajectory. It clearly highlights crucial waypoints and key actions such as \mbox{"Forward"} and \mbox{"Turn Left"}.

At the high level, the MacroPlanner takes as input the navigation map features \(\mathbf{m}\), the UAV pose \(\mathbf{s}_t=(\mathbf{p}_t,\theta_t)\) (where \(\mathbf{p}_t=(x_t,y_t,z_t)\) denotes position and \(\theta_t\) is the yaw angle at time step \(t\)), and the target description \(\mathbf{d}\). The map features \(\mathbf{m}\) include landmark information to guide the UAV toward the destination. Concretely, \(\mathbf{d}\) is a natural-language phrase (e.g., "the red car in front of the library"), and \(\mathbf{m}\) is a top-down grid of size \(C \times H \times W\). The planner outputs the next navigation sub-goal (waypoint)~\(\mathbf{w}_{k+1}\). By fusing map encodings, pose features, and the target description, the MacroPlanner~\(\mathcal{G}(\mathbf{m}, \mathbf{s}_t, \mathbf{d})\)~ predicts the next key position that the UAV should reach. Decomposing the global navigation task into a sequence of local sub-goals~\(\{\mathbf{w}_1, \mathbf{w}_2, \dots, \mathbf{w}_K\}\) (where \(\mathbf{g}\) denotes the goal position and \(\mathbf{w}_K \approx \mathbf{g}\)) helps the model avoid local optima, thereby improving overall path quality and efficiency. The MacroPlanner is triggered when the current sub-goal~\(\mathbf{w}_k\)~is achieved (i.e., when \(\|\mathbf{p}_t - \mathbf{w}_k\|_2 < \epsilon\), where \(\epsilon\)
is a threshold), and then generates the next sub-goal \(\mathbf{w}_{k+1}\).

At the low level, the MicroActor handles fine-grained action selection. It receives the current RGB observation~\(\mathbf{o}_t\), the UAV pose~\(\mathbf{s}_t\), and the current sub-goal~\(\mathbf{w}_k\)~from the MacroPlanner. It then outputs the optimal action~\(a_t^*\)~from a six-action discrete action space ~\(\mathcal{A} = \{a_{\text{go\_up}}, a_{\text{go\_down}}, a_{\text{forward}}, a_{\text{turn\_left}}, a_{\text{turn\_right}}, a_{\text{stop}}\}\). This module is modeled as a policy function~\(\pi_{\text{micro}}(\mathbf{o}_t, \mathbf{s}_t, \mathbf{w}_k)\), enabling flexible adjustment of action strategies based on local environmental perception and global planning information for adaptive navigation in complex scenarios. The MicroActor continuously selects action sequences~\(\{a_t\}_{t=0}^{T-1}\), driving state transitions~\(\mathbf{s}_t \rightarrow \mathbf{s}_{t+1}\), until the current sub-goal~\(\mathbf{w}_k\)~is achieved. At this point, the MacroPlanner is triggered to provide the next sub-goal~\(\mathbf{w}_{k+1}\). The navigation process is formalized through sequential milestones:
\[
\underbrace{\mathbf{s}_0 \xrightarrow{a_0} \mathbf{s}_1 \xrightarrow{a_1} \cdots}_{\mathclap{\text{Actions by } \pi_{\text{micro}}(\mathbf{o}_t,\mathbf{s}_t,\mathbf{w}_k)}} 
\xrightarrow{} \mathbf{w}_1 \quad \xrightarrow{} \mathbf{l}_1 \quad \cdots \quad \xrightarrow{} \mathbf{w}_K \approx \mathbf{g}
\]
Here, key landmarks $\mathcal{L} = \{\mathbf{l}_1,\dots,\mathbf{l}_M\}$ serve as spatial guidance anchors. The MacroPlanner generates sub-goals $\{\mathbf{w}_k\}_{k=1}^K$ to direct sequential progression toward $\mathbf{g}$ using landmarks in $\mathcal{L}$ as navigation references.

In this section, HTNav is first pre-trained via IL on expert trajectories to obtain well-initialized parameters and robust feature representations. Subsequently, RL with the PPO algorithm is employed to further optimize the policy through interaction with the environment.

\subsection{Map Representation Learning Module}

\begin{figure}[h]

\centering
    \includegraphics[width=9cm,height=4.0cm]{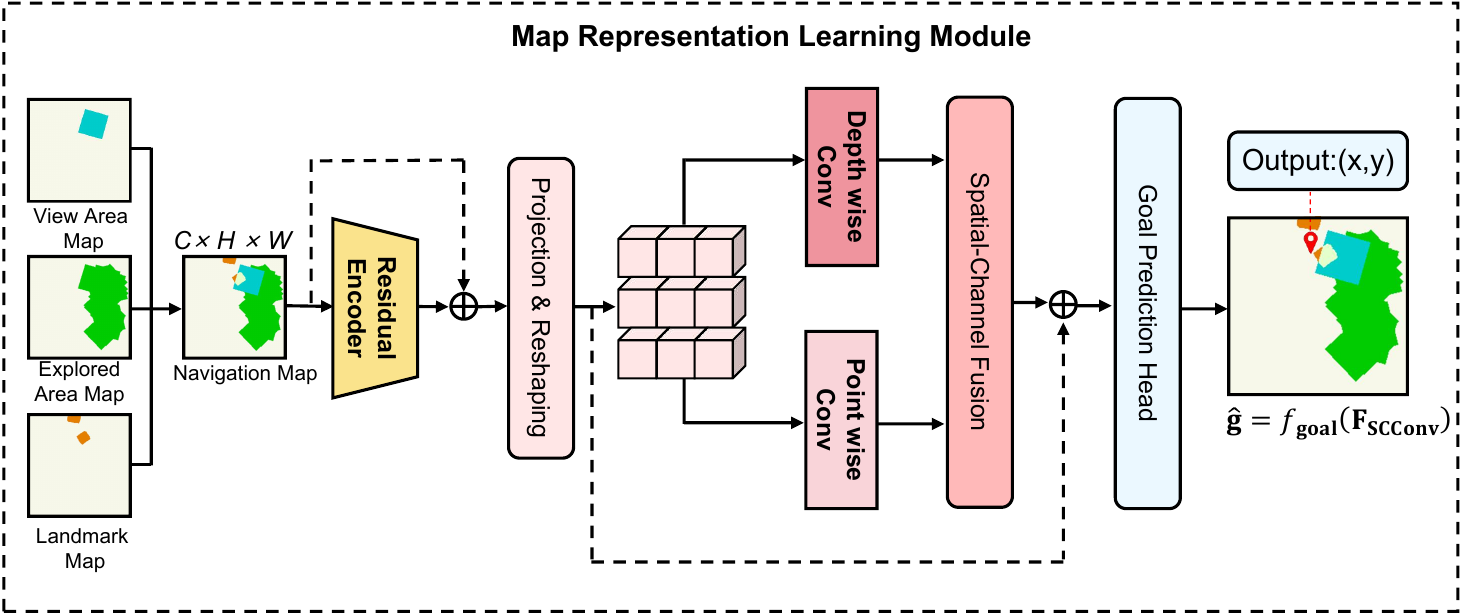}
\caption{
Schematic of the map representation learning module that integrates multi-source map inputs with a residual encoder and spatial-channel compression for goal coordinate regression.
}
\label{map}
\end{figure}

In aerial VLN tasks, prior studies leverage spatial priors, such as semantic maps~\cite{2024citynav} and semantic matrices~\cite{stmr}, as auxiliary inputs to enhance the spatial reasoning capability of navigation policies. Thus, efficient exploitation of map information is critical for UAV path planning and target localization. However, current approaches still fall short in comprehensively modeling and utilizing map information, which motivates a dedicated map learning module.

\begin{table*}[t] 
\caption{Performance comparison of methods on CityNav dataset (* indicates results obtained using the revised dataset).}
\centering
\resizebox{0.93\linewidth}{2.8cm}
{ 
\begin{tabular}{@{}lcccccccccccc@{}}
\toprule
\multirow{2}{*}{Method} & \multicolumn{4}{c}{Validation Seen} & \multicolumn{4}{c}{Validation Unseen} & \multicolumn{4}{c}{Test Unseen} \\
\cmidrule(lr){2-5} \cmidrule(lr){6-9} \cmidrule(lr){10-13}
 & NE$\downarrow$ & SR$\uparrow$ & OSR$\uparrow$ & SPL$\uparrow$ & NE$\downarrow$ & SR$\uparrow$ & OSR$\uparrow$ & SPL$\uparrow$ & NE$\downarrow$ & SR$\uparrow$ & OSR$\uparrow$ & SPL$\uparrow$ \\
\midrule
Random & 222.3 & 0.00 & 1.15 & 0.00 & 223.0 & 0.00 & 0.90 & 0.00 & 208.8 & 0.00 & 1.44 & 0.00 \\
Seq2Seq+GSM & 58.5 & 8.43 & 17.31 & 7.28 & 78.6 & 5.13 & 10.90 & 4.65 & 98.1 & 3.81 & 13.82 & 2.79 \\
CMA+GSM & 68.0 & 6.25 & 13.28 & 5.40 & 75.9 & 4.38 & 9.29 & 3.90 & 94.6 & 4.68 & 12.01 & 4.05 \\
MGP & 59.7 & 8.69 & 35.51 & 8.28 & 75.1 & 5.84 & 22.19 & 5.56 & 93.8 & 6.38 & 26.04 & 6.08 \\
MGP* & 54.1 & 10.96 & 40.82 & 10.42 & 65.6 & 8.33 & 26.75 & 7.64 & 82.6 & 9.70 & 31.46 & 8.22 \\
AerialVLN+GSM & 56.6 & 10.16 & 22.20 & 7.89 & 72.7 & 6.35 & 15.24 & 5.06 & 85.1 & 6.72 & 18.21 & 5.16 \\
FlightGPT & 66.1 & 17.57 & 30.26 & 15.78 & 68.1 & 14.69 & 29.33 & 13.24 & 76.2 & 21.20 & 35.38 & 19.24 \\
FlightGPT* & 62.7 & 19.95 & 33.33 & 18.29 & 62.4 & 16.25 & 33.87 & 14.33 & 61.4 & 24.47 & 42.17  & 21.29 \\
HTNav & 47.2 & 28.30 &  47.92 & 25.07 & 59.8 & 15.85 & 34.43 & 13.55 & 68.5 & 22.23 & 41.02 & 19.63 \\
HTNav* & \textbf{37.5} & \textbf{31.05} & \textbf{51.82} & \textbf{27.48} & \textbf{49.8} & \textbf{17.69} & \textbf{37.82} & \textbf{15.44} & \textbf{40.3} & \textbf{25.49} & \textbf{48.82} & \textbf{21.56} \\
\midrule
Human & 9.1 & 89.31 & 96.40 & 60.17 & 9.4 & 88.39 & 95.54 & 62.66 & 9.8 & 87.86 & 95.29 & 57.04 \\
\bottomrule

\end{tabular}
\label{tab:results}}
\end{table*}


Motivated by the preceding discussion, we propose a map representation learning module, as illustrated in Figure~\ref{map}. By incorporating residual connections into multi-layer convolutional feature extraction, this module alleviates vanishing gradients and preserves the original spatial information~\cite{resnet}. Each residual block performs the following computation:
\begin{equation}
\mathbf{F}^{(l+1)} = \sigma(\mathrm{BN}(\mathrm{Conv}(\mathbf{F}^{(l)})) + \mathbf{F}^{(l)}).
\end{equation}

Here, $\mathbf{F}^{(l)}$ denotes the feature map at layer $l$, and $\mathrm{Conv}$, $\mathrm{BN}$, and $\sigma$ represent convolution, batch normalization, and activation operations, respectively.

Furthermore, to better preserve spatial cues in the goal prediction head, we fuse spatial and channel features using a projection-aligned depthwise–pointwise SCConv module. SCConv jointly models correlations across spatial and channel dimensions~\cite{scconv}, reducing redundancy and improving feature utilization compared to traditional convolutions~\cite{scconv,delving}. Specifically, encoded map features are linearly transformed and fused via SCConv:
\begin{equation}
\mathbf{F}_{\text{SCConv}} = \mathrm{ReLU}\left(\mathrm{BN}\left(\mathbf{U} \odot \mathbf{C}\right)\right).
\end{equation}
Here, $\mathbf{U}$ denotes spatial convolution features and $\mathbf{C}$ represents channel fusion features. Finally, the goal prediction head processes these features to output the goal location.
\section{Experiment}
\subsection{Experimental Setup}

\noindent\textbf{Dataset.} The experiments are conducted on the CityNav dataset \cite{citynav}, which is specifically designed for aerial VLN tasks in urban environments \cite{sensaturban, cityx}. CityNav is built upon the CityRefer \cite{cityrefer} dataset and is augmented with additional target objects and navigation trajectory annotations. The CityNav dataset contains 5,850 target objects and 32,637 instruction-trajectory pairs, and it is divided into four subsets: the Training set, the Validation Seen set, the Validation Unseen set, and the Test Unseen set. In addition, similar to \cite{openfly}, tasks are classified into three difficulty levels (Easy, Medium, and Hard) based on the straight-line distance from the starting point to the target.

Our investigation of the CityNav dataset reveals missing landmark annotations in some trajectories. To address this issue, we perform a manual review of the dataset. To ensure data integrity, the correction work focuses on the following three specific types of errors, with a total of approximately 800 corrections made: (1) For cases where descriptions contain unmarked landmarks but the landmark fields are empty, complete landmark information is supplemented; (2) For incorrectly extracted landmarks from descriptions, re-extraction is performed to supplement correct landmarks; (3) For misspelled landmark names, spelling is corrected to supplement accurate names. Additionally, we remove 311 trajectories without landmark descriptions (i.e., non-navigable cases), leaving 32,326 trajectories in total. We re-run all experiments on the revised dataset. The results in Table \ref{tab:results} show that the performance of the MGP method is improved in different scenarios, which verifies the effectiveness of our data auditing and correction work.

\begin{table*}[htbp] 
\caption{ Results on Test-Unseen Set across different difficulty levels.}
\centering
{ 

\begin{tabular}{@{}lcccccccccccc@{}}
\toprule
\multirow{2}{*}{Method} & \multicolumn{4}{c}{Easy} & \multicolumn{4}{c}{Medium} & \multicolumn{4}{c}{Hard} \\
\cmidrule(lr){2-5} \cmidrule(lr){6-9} \cmidrule(lr){10-13}
 & NE$\downarrow$ & SR$\uparrow$ & OSR$\uparrow$ & SPL$\uparrow$ & NE$\downarrow$ & SR$\uparrow$ & OSR$\uparrow$ & SPL$\uparrow$ & NE$\downarrow$ & SR$\uparrow$ & OSR$\uparrow$ & SPL$\uparrow$ \\
\midrule
Random & 127.5 & 0.00 & 3.60 & 0.00 & 
212.0 & 0.00 & 0.00 & 0.00 & 
319.8 & 0.00 & 0.00 & 0.00 \\
Seq2Seq & 238.8 & 3.07 & 14.70 & 2.64 & 
246.5 & 0.43 & 3.87 & 0.38 & 
253.1 & 0.48 & 4.38 & 0.44 \\
CMA & 260.7 & 0.49 & 16.69 & 0.44 & 
241.2 & 1.10 & 7.67 & 1.09 & 
253.8 & 0.96 & 1.64 & 0.95 \\
MGP & 98.9 & 6.15 & 39.89 & 5.48 & 
90.9 & 6.29 & 21.47 & 6.21 & 
90.0 & 6.80 & 12.10 & 6.78 \\

HTNav & 
68.9 & 19.99 & 45.18 & 14.94 & 
68.2& 20.40 & 39.44& 18.69 & 
68.1 & 21.91 & 34.14& 21.11 \\
HTNav* & 
\textbf{41.1} & \textbf{23.62} & \textbf{53.82} & \textbf{17.84} & 
\textbf{40.8} & \textbf{26.26} & \textbf{48.70} & \textbf{22.42} & 
\textbf{39.9} & \textbf{28.45} & \textbf{45.19} & \textbf{27.42} \\
\bottomrule

\end{tabular}
} 

\label{tab:difficult}
\end{table*}

\noindent\textbf{Evaluation Metrics.}
We evaluate navigation performance using four standard VLN metrics \cite{r2r,vlnce,aerialvln,citynav}: navigation error (NE), success rate (SR), oracle success rate (OSR), and success weighted by path length (SPL). NE measures the Euclidean distance (in meters) between the final position of the UAV and the target, indicating basic navigation accuracy. SR denotes task success when the UAV stops within 20 m of the target. OSR assesses whether the UAV ever enters the 20 m range of the target during navigation. SPL combines success and path efficiency by comparing actual and shortest paths, indicating task efficiency.
\begin{table*}[htbp]
\caption{Ablation study: A = Tiered structure, B = Residual map encoder, C = SCConv module(results are on the revised dataset).}
\centering
{
\begin{tabular}{@{}ccccccccccccccc@{}}  
\toprule
\multirow{2}{*}{A} & \multirow{2}{*}{B} & \multirow{2}{*}{C} & \multicolumn{4}{c}{Validation Seen} & \multicolumn{4}{c}{Validation Unseen} & \multicolumn{4}{c}{Test Unseen} \\
\cmidrule(lr){4-7} \cmidrule(lr){8-11} \cmidrule(lr){12-15}
& & & NE$\downarrow$ & SR$\uparrow$ & OSR$\uparrow$ & SPL$\uparrow$ & NE$\downarrow$ & SR$\uparrow$ & OSR$\uparrow$ & SPL$\uparrow$ & NE$\downarrow$ & SR$\uparrow$ & OSR$\uparrow$ & SPL$\uparrow$ \\
\midrule
 & MGP(baseline)& & 54.1 & 10.96 & 40.82 & 10.42 & 65.6 & 8.33 & 26.75 & 7.64 & 82.6 & 9.70 & 31.46 & 8.22 \\
 & MGP+IL-RL &  & 47.1 & 23.56 & 45.55 & 20.03 & 58.3 & 12.83 & 34.07 & 10.18 & 47.8 & 18.95 & 43.08  & 14.82 \\  
$\checkmark$ & - & - & 40.7 & 26.15 & 49.03 & 24.81 & 52.1 & 15.09 & 33.15 & 14.19 & 44.0 & 24.03 & 44.99 & 20.26 \\  
$\checkmark$ & $\checkmark$ &- & 39.8 & 28.07 & 49.45 & 24.72 & 51.2 & 16.05 & 37.19 & 14.09 & 41.0 & 24.18 & 46.53 & 21.02 \\  
$\checkmark$ & - & $\checkmark$ & 40.5 & 27.77 & 49.35 & 24.68 & 51.4 & 15.72 & 35.52 & 14.16 & 41.8 & 24.16 & 47.45 & 21.03 \\  

$\checkmark$ & $\checkmark$ & $\checkmark$ & \textbf{37.5} & \textbf{31.05} & \textbf{51.82} & \textbf{27.48} & \textbf{49.8} & \textbf{17.69} & \textbf{37.82} & \textbf{15.44} & \textbf{40.3} & \textbf{25.49} & \textbf{48.82} & \textbf{21.56} \\  
\bottomrule
\end{tabular}
}
\label{tab:ablation}
\end{table*}

\noindent\textbf{Comparison Methods.}
Our selected baselines range from a Random policy and Seq2Seq+GSM model \cite{r2r,citynav} to stronger approaches such as MGP \cite{2024citynav}, CMA+GSM \cite{aerialvln,citynav} and AerialVLN+GSM \cite{aerialvln,citynav}. Here, GSM stands for the Geographic Semantic Map proposed in \cite{citynav}. Finally, we benchmark HTNav against FlightGPT, a state-of-the-art model that leverages large language models.

\noindent\textbf{Implementation and Training Details.} All of our experiments are conducted on an NVIDIA RTX A5000 GPU. The imitation learning model is trained using the AdamW optimizer \cite{adamw} with a batch size of 4 and a learning rate of 1.5e-3. During the RL stage, the policy is optimized with PPO using a learning rate of 3e-5. We use ResNet-50 backbones as the visual and depth encoders, which are pre-trained on ImageNet \cite{imagenet} and PointGoalNav \cite{PointGoalNav}, respectively.

\subsection{Experimental Results}
\noindent\textbf{Overall Performance.}
To ensure fairness, we conduct experiments on both the original and revised datasets. As shown in Table \ref{tab:results}, HTNav achieves the best performance in both cases. Compared to the baseline MGP approach, HTNav improves across all metrics, with the success rate more than doubling, from 9.70\% to 25.49\% (Test Unseen). Notably, human performance (measured via human evaluation) still significantly outperforms current agents, highlighting a large gap in spatial reasoning and dynamic decision-making. This suggests that future models require further improvements to approach human-level navigation.

\noindent\textbf{Quantitative Results.} Experimental results indicate that HTNav offers significant advantages in both navigation accuracy and efficiency (Test Unseen split: NE = 40.3 m, SR = 25.49\%). HTNav achieves substantially lower NE values in all scenarios compared to existing baselines, indicating that the navigation agent can reach locations closer to the target. Furthermore, HTNav consistently achieves the highest OSR, suggesting the agent approaches the target more frequently during navigation, which increases the likelihood of successful task completion. Our method also achieves significantly higher SPL than other baselines, demonstrating better path efficiency without sacrificing success rates.
\begin{figure*}[!h]

\begin{center}

\includegraphics[width=18cm,height=5cm]{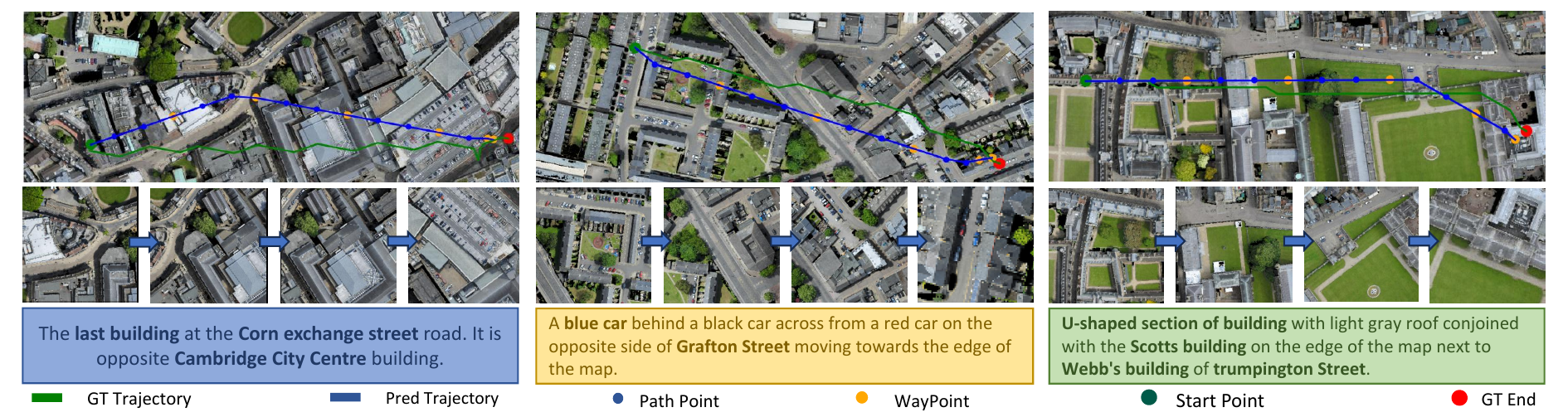}

\end{center}

\caption{Visualization of navigation trajectories. The left panel illustrates the long-path navigation capability, the middle panel demonstrates the small-target localization ability, and the right panel exhibits the efficiency of the navigation path. }

\label{figure3}

\end{figure*}

\noindent\textbf{Performance in Different Task Difficulties.}
As shown in Table~\ref{tab:difficult}, HTNav demonstrates outstanding performance across all difficulty levels, with a particularly significant improvement in success rate when faced with challenging scenarios. It is worth noting that, based on our analysis of the dataset, a minority of tasks—particularly in the easy and medium difficulty tiers—have shorter instructions. Shorter instructions can make target descriptions more ambiguous or insufficiently specific. As a result, the success rate on these tasks tends to be lower. This phenomenon is also observed for the MGP method.


\noindent\textbf{Ablation Study.}
In this section, we conduct ablation studies on HTNav using the revised dataset, with the results summarized in Table \ref{tab:ablation}. Specifically, we progressively introduce the IL-RL framework, the tiered structure, the residual map encoder, and the SCConv module, and evaluate the model's navigation performance and generalization across environments under each setting. 
Experimental results show that adopting the IL-RL framework significantly improves most metrics (Test-Unseen SR: 9.70\%→18.95\%). On this basis, incorporating the tiered structure leads to a reduction in NE and an improvement in SPL, validating its effectiveness. Integrating the residual map encoder and the SCConv module boosts performance across all scenarios, with particularly strong gains in seen environments. This suggests that improving success rates in unseen environments still relies primarily on reinforcement learning.

\noindent\textbf{Ablation Study under Different RL Weight.}
Table~\ref{tab:rl_weights} summarizes our investigation of the reinforcement learning weight $\lambda_{\text{RL}}$. The model with $\lambda_{\text{RL}}=0$ (i.e., without RL) shows the weakest performance. As $\lambda_{\text{RL}}$ increases from 0.10 to 0.20, all metrics steadily improve (e.g., SR rises from 21.89\% to 25.49\%). The model performs best at $\lambda_{\text{RL}}=0.20$, achieving the highest SR (25.49\%). However, increasing the weight further to 0.25 and 0.30 leads to degraded performance across all metrics, suggesting that an excessively large $\lambda_{\text{RL}}$ may introduce training instability.


\begin{table}[!h]
\centering
\caption{Performance metrics under different RL weights.}
\begin{tabular}{lrrrr} 
\toprule
\multirow{2}{*}{$\lambda_{\text{RL}}$} & \multicolumn{4}{c}{Test-Unseen} \\ 
\cmidrule(lr){2-5} 
& {NE$\downarrow$} & {SR$\uparrow$} & {OSR$\uparrow$} & {SPL$\uparrow$} \\ 
\midrule
0(w/o RL) & 53.0 & 18.90 & 40.28 & 17.53 \\
0.10 & 44.0 & 21.89 & 44.16 & 19.88 \\
0.15 &  43.6 & 24.16 &  48.02 & 21.03 \\
0.20 & \textbf {40.3} & \textbf {25.49} &  \textbf{48.82} & \textbf{21.56} \\
0.25 & 41.9 & 24.29 & 47.42 & 20.89 \\
0.30 & 42.0 & 23.56  & 42.51 & 19.73 \\
\bottomrule
\end{tabular}
\label{tab:rl_weights}
\end{table}
\noindent\textbf{Ablation Study on Navigation Submaps.}
Our method aims to balance performance and computational efficiency. While the original MGP framework is effective, using five distinct submaps introduces considerable complexity. We hypothesize that a more streamlined configuration could achieve comparable performance with fewer resources. The ablation study in Table~\ref{tab:ablation_mgp} shows that the model preserves strong performance after removing the target and surroundings maps, supporting our simplified map design. In contrast, discarding the landmark\_map causes a significant drop in SR, indicating its critical importance. This leaner design eliminates the need for target-level semantic segmentation, thereby avoiding GroundingDINO~\cite{liu2024grounding} and Mobile-SAM~\cite{zhang2023faster} while simplifying the overall pipeline.

\begin{table}[htbp]
  \centering
    \caption{Effectiveness evaluation of Target and Surroundings Map.}
    
    \begin{tabular}{lc}
    \toprule
    Method & \multicolumn{1}{c}{SR $\uparrow$} \\
    \midrule
    MGP \cite{2024citynav}  & 5.84 \\
    w/o target \& surroundings maps & 5.81 \\
    w/o landmark map & 0.47 \\
    HTNav & 28.30 \\
    w/ target \& surroundings maps & 28.62\\
    w/o landmark map & 1.86 \\
    \bottomrule
    \end{tabular}%
    
  \label{tab:ablation_mgp}
\end{table}



\noindent\textbf{Qualitative Results.} Figure \ref{figure3} presents three successful navigation cases of the HTNav. The figure compares the predicted trajectories with the ground-truth trajectories: the left panel illustrates a navigation route that is structurally analogous to the one in Figure~\ref{figure1}(c), thereby demonstrating stable performance in long-range navigation; the central panel exhibits the system's precise localization and approach to a small target (the blue car); the right panel further reveals close alignment between predicted and ground-truth trajectories, highlighting improved navigation performance in complex environments.
\section{Conclusion}
In aerial VLN, UAVs in complex urban environments face three core challenges: limited generalization to unfamiliar scenes, suboptimal long-range planning, and insufficient understanding of spatial continuity. In this paper, we propose HTNav, a hybrid IL–RL navigation framework with a tiered decision-making mechanism that decomposes long-range navigation into global waypoint generation and reactive local action selection, together with a map representation learning module that enhances multi-step spatial reasoning. These innovations significantly improve a UAV’s global planning capability in long-range scenarios. Our model achieves state-of-the-art performance on both the original and revised CityNav benchmarks, validating the effectiveness of HTNav on complex tasks and also confirming the significant impact of dataset revisions.


{
    \small
    \bibliographystyle{ieeenat_fullname}
    \bibliography{main}

\begin{thebibliography}{42}
\providecommand{\natexlab}[1]{#1}
\providecommand{\url}[1]{\texttt{#1}}
\expandafter\ifx\csname urlstyle\endcsname\relax
  \providecommand{\doi}[1]{doi: #1}\else
  \providecommand{\doi}{doi: \begingroup \urlstyle{rm}\Url}\fi

\bibitem[Anderson et~al.(2018{\natexlab{a}})Anderson, Chang, Chaplot, Dosovitskiy, Gupta, Koltun, Kosecka, Malik, Mottaghi, Savva, et~al.]{PointGoalNav}
Peter Anderson, Angel Chang, Devendra~Singh Chaplot, Alexey Dosovitskiy, Saurabh Gupta, Vladlen Koltun, Jana Kosecka, Jitendra Malik, Roozbeh Mottaghi, Manolis Savva, et~al.
\newblock On evaluation of embodied navigation agents.
\newblock \emph{arXiv preprint arXiv:1807.06757}, 2018{\natexlab{a}}.

\bibitem[Anderson et~al.(2018{\natexlab{b}})Anderson, Wu, Teney, Bruce, Johnson, S{\"u}nderhauf, Reid, Gould, and Van Den~Hengel]{r2r}
Peter Anderson, Qi Wu, Damien Teney, Jake Bruce, Mark Johnson, Niko S{\"u}nderhauf, Ian Reid, Stephen Gould, and Anton Van Den~Hengel.
\newblock Vision-and-language navigation: Interpreting visually-grounded navigation instructions in real environments.
\newblock In \emph{Proceedings of the IEEE Conference on Computer Vision and Pattern Recognition}, pages 3674--3683, 2018{\natexlab{b}}.

\bibitem[Bu et~al.(2021)Bu, Peng, Yan, Tan, and Zhang]{gaia}
Xingyuan Bu, Junran Peng, Junjie Yan, Tieniu Tan, and Zhaoxiang Zhang.
\newblock Gaia: A transfer learning system of object detection that fits your needs.
\newblock In \emph{Proceedings of the IEEE/CVF Conference on Computer Vision and Pattern Recognition}, pages 274--283, 2021.

\bibitem[Cai et~al.(2025{\natexlab{a}})Cai, Dong, Rao, Deng, Tan, Chen, Wang, Wang, Huang, Sumalee, et~al.]{SA-GCS}
Hengxing Cai, Jinhan Dong, Yijie Rao, Jingcheng Deng, Jingjun Tan, Qien Chen, Haidong Wang, Zhen Wang, Shiyu Huang, Agachai Sumalee, et~al.
\newblock Sa-gcs: Semantic-aware gaussian curriculum scheduling for uav vision-language navigation.
\newblock \emph{arXiv preprint arXiv:2508.00390}, 2025{\natexlab{a}}.

\bibitem[Cai et~al.(2025{\natexlab{b}})Cai, Dong, Tan, Deng, Li, Gao, Wang, Su, Sumalee, and Zhong]{flightgpt}
Hengxing Cai, Jinhan Dong, Jingjun Tan, Jingcheng Deng, Sihang Li, Zhifeng Gao, Haidong Wang, Zicheng Su, Agachai Sumalee, and Renxin Zhong.
\newblock {F}light{GPT}: Towards generalizable and interpretable {UAV} vision-and-language navigation with vision-language models.
\newblock In \emph{Proceedings of the 2025 Conference on Empirical Methods in Natural Language Processing}, pages 6670--6687. Association for Computational Linguistics, 2025{\natexlab{b}}.

\bibitem[Chen et~al.(2019)Chen, Suhr, Misra, Snavely, and Artzi]{touchdown}
Howard Chen, Alane Suhr, Dipendra Misra, Noah Snavely, and Yoav Artzi.
\newblock Touchdown: Natural language navigation and spatial reasoning in visual street environments.
\newblock In \emph{Proceedings of the IEEE/CVF Conference on Computer Vision and Pattern Recognition}, pages 12538--12547, 2019.

\bibitem[Das et~al.(2018)Das, Datta, Gkioxari, Lee, Parikh, and Batra]{embodiedqa}
Abhishek Das, Samyak Datta, Georgia Gkioxari, Stefan Lee, Devi Parikh, and Dhruv Batra.
\newblock Embodied question answering.
\newblock In \emph{Proceedings of the IEEE Conference on Computer Vision and Pattern Recognition}, pages 1--10, 2018.

\bibitem[Deng et~al.(2009)Deng, Dong, Socher, Li, Li, and Fei-Fei]{imagenet}
Jia Deng, Wei Dong, Richard Socher, Li-Jia Li, Kai Li, and Li Fei-Fei.
\newblock Imagenet: A large-scale hierarchical image database.
\newblock In \emph{2009 IEEE conference on Computer Vision and Pattern Recognition}, pages 248--255. Ieee, 2009.

\bibitem[Fan et~al.(2023)Fan, Chen, Jiang, Zhou, Zhang, and Wang]{c:avdn}
Yue Fan, Winson Chen, Tongzhou Jiang, Chun Zhou, Yi Zhang, and Xin Wang.
\newblock Aerial vision-and-dialog navigation.
\newblock In \emph{Findings of the Association for Computational Linguistics: ACL 2023}, pages 3043--3061, 2023.

\bibitem[Gao et~al.(2024)Gao, Wang, Jing, Wang, Li, and Zhao]{stmr}
Yunpeng Gao, Zhigang Wang, Linglin Jing, Dong Wang, Xuelong Li, and Bin Zhao.
\newblock Aerial vision-and-language navigation via semantic-topo-metric representation guided llm reasoning.
\newblock \emph{arXiv preprint arXiv:2410.08500}, 2024.

\bibitem[Gao et~al.(2025)Gao, Li, You, Liu, Li, Chen, Chen, Tang, Wang, Yang, et~al.]{openfly}
Yunpeng Gao, Chenhui Li, Zhongrui You, Junli Liu, Zhen Li, Pengan Chen, Qizhi Chen, Zhonghan Tang, Liansheng Wang, Penghui Yang, et~al.
\newblock Openfly: A comprehensive platform for aerial vision-language navigation.
\newblock \emph{arXiv preprint arXiv:2502.18041}, 2025.

\bibitem[He et~al.(2016)He, Zhang, Ren, and Sun]{resnet}
Kaiming He, Xiangyu Zhang, Shaoqing Ren, and Jian Sun.
\newblock Deep residual learning for image recognition.
\newblock In \emph{Proceedings of the IEEE Conference on Computer Vision and Pattern Recognition}, pages 770--778, 2016.

\bibitem[He et~al.(2021)He, Huang, Wu, Yang, An, Sima, and Wang]{landmarkrxr}
Keji He, Yan Huang, Qi Wu, Jianhua Yang, Dong An, Shuanglin Sima, and Liang Wang.
\newblock {Landmark-RxR}: Solving vision-and-language navigation with fine-grained alignment supervision.
\newblock \emph{Advances in Neural Information Processing Systems}, 34:\penalty0 652--663, 2021.

\bibitem[He et~al.(2023)He, Si, Lu, Huang, Wang, and Wang]{frequency}
Keji He, Chenyang Si, Zhihe Lu, Yan Huang, Liang Wang, and Xinchao Wang.
\newblock Frequency-enhanced data augmentation for vision-and-language navigation.
\newblock \emph{Advances in Neural Information Processing Systems}, 36:\penalty0 4351--4364, 2023.

\bibitem[Hu et~al.(2022)Hu, Yang, Khalid, Xiao, Trigoni, and Markham]{sensaturban}
Qingyong Hu, Bo Yang, Sheikh Khalid, Wen Xiao, Niki Trigoni, and Andrew Markham.
\newblock Sensaturban: Learning semantics from urban-scale photogrammetric point clouds.
\newblock \emph{International Journal of Computer Vision}, 130\penalty0 (2):\penalty0 316--343, 2022.

\bibitem[Jain et~al.(2019)Jain, Magalhaes, Ku, Vaswani, Ie, and Baldridge]{r4r}
Vihan Jain, Gabriel Magalhaes, Alexander Ku, Ashish Vaswani, Eugene Ie, and Jason Baldridge.
\newblock Stay on the path: Instruction fidelity in vision-and-language navigation.
\newblock In \emph{Proceedings of the 57th Annual Meeting of the Association for Computational Linguistics}, pages 1862--1872, 2019.

\bibitem[Krantz et~al.(2020)Krantz, Wijmans, Majumdar, Batra, and Lee]{vlnce}
Jacob Krantz, Erik Wijmans, Arjun Majumdar, Dhruv Batra, and Stefan Lee.
\newblock Beyond the nav-graph: Vision-and-language navigation in continuous environments.
\newblock In \emph{European Conference on Computer Vision}, pages 104--120. Springer, 2020.

\bibitem[Ku et~al.(2020)Ku, Anderson, Patel, Ie, and Baldridge]{rxr}
Alexander Ku, Peter Anderson, Roma Patel, Eugene Ie, and Jason Baldridge.
\newblock Room-across-room: Multilingual vision-and-language navigation with dense spatiotemporal grounding.
\newblock In \emph{Proceedings of the 2020 Conference on Empirical Methods in Natural Language Processing}, pages 4392--4412, 2020.

\bibitem[Lee et~al.(2024)Lee, Miyanishi, Kurita, Sakamoto, Azuma, Matsuo, and Inoue]{2024citynav}
Jungdae Lee, Taiki Miyanishi, Shuhei Kurita, Koya Sakamoto, Daichi Azuma, Yutaka Matsuo, and Nakamasa Inoue.
\newblock {CityNav}: Language-goal aerial navigation dataset with geographic information.
\newblock \emph{arXiv preprint arXiv:2406.14240}, 2024.

\bibitem[Lee et~al.(2025)Lee, Miyanishi, Kurita, Sakamoto, Azuma, Matsuo, and Inoue]{citynav}
Jungdae Lee, Taiki Miyanishi, Shuhei Kurita, Koya Sakamoto, Daichi Azuma, Yutaka Matsuo, and Nakamasa Inoue.
\newblock {CityNav}: A large-scale dataset for real-world aerial navigation.
\newblock In \emph{Proceedings of the IEEE/CVF International Conference on Computer Vision}, pages 5912--5922, 2025.

\bibitem[Li et~al.(2023)Li, Wen, and He]{scconv}
Jiafeng Li, Ying Wen, and Lianghua He.
\newblock {SCConv}: Spatial and channel reconstruction convolution for feature redundancy.
\newblock In \emph{Proceedings of the IEEE/CVF Conference on Computer Vision and Pattern Recognition}, pages 6153--6162, 2023.

\bibitem[Liu et~al.(2023)Liu, Zhang, Qi, Wang, Zhang, and Wu]{aerialvln}
Shubo Liu, Hongsheng Zhang, Yuankai Qi, Peng Wang, Yanning Zhang, and Qi Wu.
\newblock {AerialVLN}: Vision-and-language navigation for uavs.
\newblock In \emph{Proceedings of the IEEE/CVF International Conference on Computer Vision}, pages 15384--15394, 2023.

\bibitem[Liu et~al.(2024{\natexlab{a}})Liu, Zeng, Ren, Li, Zhang, Yang, Jiang, Li, Yang, Su, et~al.]{liu2024grounding}
Shilong Liu, Zhaoyang Zeng, Tianhe Ren, Feng Li, Hao Zhang, Jie Yang, Qing Jiang, Chunyuan Li, Jianwei Yang, Hang Su, et~al.
\newblock {Grounding DINO}: Marrying dino with grounded pre-training for open-set object detection.
\newblock In \emph{European Conference on Computer Vision}, pages 38--55. Springer, 2024{\natexlab{a}}.

\bibitem[Liu et~al.(2024{\natexlab{b}})Liu, Yao, Yue, Xu, Sun, and Fu]{navagent}
Youzhi Liu, Fanglong Yao, Yuanchang Yue, Guangluan Xu, Xian Sun, and Kun Fu.
\newblock Navagent: Multi-scale urban street view fusion for uav embodied vision-and-language navigation.
\newblock \emph{arXiv preprint arXiv:2411.08579}, 2024{\natexlab{b}}.

\bibitem[Loshchilov et~al.(2017)Loshchilov, Hutter, et~al.]{adamw}
Ilya Loshchilov, Frank Hutter, et~al.
\newblock Fixing weight decay regularization in adam.
\newblock \emph{arXiv preprint arXiv:1711.05101}, 5\penalty0 (5):\penalty0 5, 2017.

\bibitem[Miyanishi et~al.(2023)Miyanishi, Kitamori, Kurita, Lee, Kawanabe, and Inoue]{cityrefer}
Taiki Miyanishi, Fumiya Kitamori, Shuhei Kurita, Jungdae Lee, Motoaki Kawanabe, and Nakamasa Inoue.
\newblock Cityrefer: geography-aware 3d visual grounding dataset on city-scale point cloud data.
\newblock \emph{arXiv preprint arXiv:2310.18773}, 2023.

\bibitem[Pan et~al.(2023)Pan, He, Peng, Zhang, Sui, and Zhang]{baeformer}
Cong Pan, Yonghao He, Junran Peng, Qian Zhang, Wei Sui, and Zhaoxiang Zhang.
\newblock {BAEFormer}: Bi-directional and early interaction transformers for bird's eye view semantic segmentation.
\newblock In \emph{Proceedings of the IEEE/CVF Conference on Computer Vision and Pattern Recognition}, pages 9590--9599, 2023.

\bibitem[Pan et~al.(2024)Pan, Peng, and Zhang]{depth}
Cong Pan, Junran Peng, and Zhaoxiang Zhang.
\newblock Depth-guided vision transformer with normalizing flows for monocular 3d object detection.
\newblock \emph{IEEE/CAA Journal of Automatica Sinica}, 11\penalty0 (3):\penalty0 673--689, 2024.

\bibitem[Peng et~al.(2020)Peng, Bu, Sun, Zhang, Tan, and Yan]{large-scale}
Junran Peng, Xingyuan Bu, Ming Sun, Zhaoxiang Zhang, Tieniu Tan, and Junjie Yan.
\newblock Large-scale object detection in the wild from imbalanced multi-labels.
\newblock In \emph{Proceedings of the IEEE/CVF Conference on Computer Vision and Pattern Recognition}, pages 9709--9718, 2020.

\bibitem[Peng et~al.(2023)Peng, Chang, Yin, Bu, Sun, Xie, Zhang, Tian, and Zhang]{gaiaun}
Junran Peng, Qing Chang, Haoran Yin, Xingyuan Bu, Jiajun Sun, Lingxi Xie, Xiaopeng Zhang, Qi Tian, and Zhaoxiang Zhang.
\newblock Gaia-universe: Everything is super-netify.
\newblock \emph{IEEE Transactions on Pattern Analysis and Machine Intelligence}, 45\penalty0 (10):\penalty0 11856--11868, 2023.

\bibitem[Qi et~al.(2020)Qi, Wu, Anderson, Wang, Wang, Shen, and Hengel]{reverie}
Yuankai Qi, Qi Wu, Peter Anderson, Xin Wang, William~Yang Wang, Chunhua Shen, and Anton van~den Hengel.
\newblock Reverie: Remote embodied visual referring expression in real indoor environments.
\newblock In \emph{Proceedings of the IEEE/CVF Conference on Computer Vision and Pattern Recognition}, pages 9982--9991, 2020.

\bibitem[Schulman et~al.(2017)Schulman, Wolski, Dhariwal, Radford, and Klimov]{ppo}
John Schulman, Filip Wolski, Prafulla Dhariwal, Alec Radford, and Oleg Klimov.
\newblock Proximal policy optimization algorithms.
\newblock \emph{arXiv preprint arXiv:1707.06347}, 2017.

\bibitem[Shah et~al.(2017)Shah, Dey, Lovett, and Kapoor]{airsim}
Shital Shah, Debadeepta Dey, Chris Lovett, and Ashish Kapoor.
\newblock Airsim: High-fidelity visual and physical simulation for autonomous vehicles.
\newblock In \emph{Field and service robotics: Results of the 11th international conference}, pages 621--635. Springer, 2017.

\bibitem[Shridhar et~al.(2020)Shridhar, Thomason, Gordon, Bisk, Han, Mottaghi, Zettlemoyer, and Fox]{alfred}
Mohit Shridhar, Jesse Thomason, Daniel Gordon, Yonatan Bisk, Winson Han, Roozbeh Mottaghi, Luke Zettlemoyer, and Dieter Fox.
\newblock Alfred: A benchmark for interpreting grounded instructions for everyday tasks.
\newblock In \emph{Proceedings of the IEEE/CVF Conference on Computer Vision and Pattern Recognition}, pages 10740--10749, 2020.

\bibitem[Su et~al.(2025)Su, An, Chen, Yu, Ning, Ling, Huang, and Wang]{c:fgavdn}
Yifei Su, Dong An, Kehan Chen, Weichen Yu, Baiyang Ning, Yonggen Ling, Yan Huang, and Liang Wang.
\newblock Learning fine-grained alignment for aerial vision-dialog navigation.
\newblock \emph{In Proceedings of the AAAI Conference on Artificial Intelligence}, 39\penalty0 (7):\penalty0 7060--7068, 2025.

\bibitem[Vaswani et~al.(2017)Vaswani, Shazeer, Parmar, Uszkoreit, Jones, Gomez, Kaiser, and Polosukhin]{attention}
Ashish Vaswani, Noam Shazeer, Niki Parmar, Jakob Uszkoreit, Llion Jones, Aidan~N Gomez, {\L}ukasz Kaiser, and Illia Polosukhin.
\newblock Attention is all you need.
\newblock \emph{Advances in Neural Information Processing Systems}, 30, 2017.

\bibitem[Wang et~al.(2025)Wang, Yang, Wang, Kwan, Chen, Wu, Li, Liao, and Liu]{openuav}
Xiangyu Wang, Donglin Yang, Ziqin Wang, Hohin Kwan, Jinyu Chen, Wenjun Wu, Hongsheng Li, Yue Liao, and Si Liu.
\newblock Towards realistic {UAV} vision-language navigation: Platform, benchmark, and methodology.
\newblock In \emph{The Thirteenth International Conference on Learning Representations}, 2025.

\bibitem[Xu et~al.(2025)Xu, Hu, Gao, Zhu, Zhao, Li, and Yin]{geonav}
Haotian Xu, Yue Hu, Chen Gao, Zhengqiu Zhu, Yong Zhao, Yong Li, and Quanjun Yin.
\newblock Geonav: Empowering mllms with explicit geospatial reasoning abilities for language-goal aerial navigation.
\newblock \emph{arXiv preprint arXiv:2504.09587}, 2025.

\bibitem[Yeung et~al.(2024)Yeung, Pham, Zhang, Fountaine, and Raman]{hybrid}
Christopher Yeung, Benjamin Pham, Zihan Zhang, Katherine~T Fountaine, and Aaswath~P Raman.
\newblock Hybrid supervised and reinforcement learning for the design and optimization of nanophotonic structures.
\newblock \emph{Optics Express}, 32\penalty0 (6):\penalty0 9920--9930, 2024.

\bibitem[Zhang et~al.(2023)Zhang, Han, Qiao, Kim, Bae, Lee, and Hong]{zhang2023faster}
Chaoning Zhang, Dongshen Han, Yu Qiao, Jung~Uk Kim, Sung-Ho Bae, Seungkyu Lee, and Choong~Seon Hong.
\newblock Faster segment anything: Towards lightweight sam for mobile applications.
\newblock \emph{arXiv preprint arXiv:2306.14289}, 2023.

\bibitem[Zhang et~al.(2024)Zhang, Zhou, Wang, Luo, Wang, Li, Zhang, and Peng]{cityx}
Shougao Zhang, Mengqi Zhou, Yuxi Wang, Chuanchen Luo, Rongyu Wang, Yiwei Li, Zhaoxiang Zhang, and Junran Peng.
\newblock Cityx: Controllable procedural content generation for unbounded 3d cities.
\newblock \emph{arXiv preprint arXiv:2407.17572}, 2024.

\bibitem[Zhang et~al.(2022)Zhang, Pan, and Peng]{delving}
Zhaoxiang Zhang, Cong Pan, and Junran Peng.
\newblock Delving into the effectiveness of receptive fields: Learning scale-transferrable architectures for practical object detection.
\newblock \emph{International Journal of Computer Vision}, 130\penalty0 (4):\penalty0 970--989, 2022.

\end{thebibliography}
}

\end{document}